\pdfoutput=1
\documentclass[abstract=true]{scrartcl}

\usepackage[normal, bf]{caption}  
\usepackage{booktabs} 
\usepackage{natbib}
\usepackage{algorithmicx}
\usepackage{algorithm}
\usepackage{algpseudocode}
\usepackage{subfig}
\usepackage{xcolor}
\usepackage{graphicx}
\usepackage{lineno,hyperref}

\hypersetup{
	colorlinks   = true, 
	ocgcolorlinks = false,
	urlcolor     = blue, 
	linkcolor    = blue, 
	citecolor   = red 
}

\begin{document}

\title{How to avoid machine learning pitfalls:\\a guide for academic researchers}

\author{Michael A. Lones\footnote{School of Mathematical and Computer Sciences, Heriot-Watt University, Edinburgh, Scotland, UK, Email: \href{mailto:m.lones@hw.ac.uk}{m.lones@hw.ac.uk}, Web: \href{http://www.macs.hw.ac.uk/\%7Eml355}{www.macs.hw.ac.uk/$\sim$ml355}, Substack: \href{https://fetchdecodeexecute.substack.com/}{Fetch Decode Execute}.}}

\date{}
\maketitle

\vspace{-14mm}
\begin{abstract}
\noindent Mistakes in machine learning practice are commonplace, and can result in a loss of confidence in the findings and products of machine learning. This guide outlines common mistakes that occur when using machine learning, and what can be done to avoid them. Whilst it should be accessible to anyone with a basic understanding of machine learning techniques, it focuses on issues that are of particular concern within academic research, such as the need to do rigorous comparisons and reach valid conclusions. It covers five stages of the machine learning process: what to do before model building, how to reliably build models, how to robustly evaluate models, how to compare models fairly, and how to report results.
\end{abstract}


\section{Introduction}

It's easy to make mistakes when applying machine learning (ML), and these mistakes can result in ML models that fail to work as expected when applied to data not seen during training and testing \citep{liao2021are}. This is a problem for practitioners, since it leads to the failure of ML projects. However, it is also a problem for society, since it erodes trust in the findings and products of ML \citep{gibney2022ai}.

This guide aims to help newcomers avoid some of these mistakes. It's written by an academic, and focuses on lessons learnt whilst doing ML research in academia. Whilst primarily aimed at students and scientific researchers, it should be accessible to anyone getting started in ML, and only assumes a basic knowledge of ML techniques. However, unlike similar guides aimed at a more general audience, it includes topics that are of a particular concern to academia, such as the need to rigorously evaluate and compare models in order to get work published.

To make it more readable, the guidance is written informally, in a \textit{Dos and Don'ts} style. It's not intended to be exhaustive, and references (with publicly-accessible URLs where available) are provided for further reading. Since it doesn't cover issues specific to particular academic subjects, it's recommended that readers also consult subject-specific guidance where available, e.g.\ in clinical medicine \citep{stevens2020recommendations}, genomics \citep{whalen2022navigating}, environmental research \citep{zhu2023machine},  materials science \citep{karande2022strategic}, business and marketing \citep{van2022overcoming}, computer security \citep{arp2022and} and social science \citep{malik2020hierarchy}.

The review is divided into five sections.  \nameref{before} covers issues that can occur early in the ML process, and focuses on the correct use of data and adequate consideration of the context in which ML is being applied. \nameref{build} then covers pitfalls that occur during the selection and training of models and their components. \nameref{evaluate} presents pitfalls that can lead to an incorrect understanding of model performance. \nameref{compare} then extends this to the situation where models are being compared, discussing how common pitfalls can lead to misleading findings. \nameref{report} focuses on reproducibility and factors that can lead to incomplete or deceptive reporting.

\section*{Changes}

ML pitfalls are not static, and continue to evolve as ML develops. To address this, this guide has been updated annually since it was first released in 2021, and it will continue to be updated in the future. Feedback is welcome. If you cite it, please include the arXiv version number (currently v5\footnote{This version is published as ``Avoiding machine learning pitfalls" in \textit{Patterns} (Cell Press) \citep{lones2024avoiding_ml_pitfalls}}).

\paragraph*{Changes from \href{https://arxiv.org/abs/2108.02497v4}{v4}}
Added \nameref{baselines}, \nameref{clean} and \nameref{fairness}. Extended \nameref{look}, \nameref{augmentation}, \nameref{deployment}, \nameref{leakage} and \nameref{multiple}.

\paragraph*{Changes from \href{https://arxiv.org/abs/2108.02497v3}{v3}}
Added \nameref{checklists} and \nameref{datawhere}. Rewrote \nameref{multiple}. Revised \nameref{trends}, \nameref{feature}, \nameref{validation}, \nameref{accuracy} and \nameref{ensemble}. Extended \nameref{testsets}.

\paragraph*{Changes from \href{https://arxiv.org/abs/2108.02497v2}{v2}}
Added illustrations. Added \nameref{spurious}, \nameref{temporal} and \nameref{trends}.

\paragraph*{Changes from \href{https://arxiv.org/abs/2108.02497v1}{v1}}
Added \nameref{badaugmentation} and \nameref{dnn}. Rewrote \nameref{inappropriate}. Expanded \nameref{leakage}, \nameref{significance} and \nameref{transparent}.

\newpage
\renewcommand{\baselinestretch}{0.98}\normalsize
{\small \tableofcontents}
\renewcommand{\baselinestretch}{1.00}\normalsize
\newpage

\section{Before you start to build models} \label{before}

It's normal to want to rush into training and evaluating models, but it's important to take the time to think about the goals of a project, to fully understand the data that will be used to support these goals, to consider any limitations of the data that need to be addressed, and to understand what's already been done in your field. If you don't do these things, then you may end up with results that are hard to publish, or models that are not appropriate for their intended purpose.

\subsection{Do think about how and where you will use data} \label{datawhere}
Data is central to most ML projects, but is often in short supply. Therefore it's important to think carefully about what data you need and how and where you will use it. Abstractly, you need data for two things, training models and testing models. However, for various reasons, this does not necessarily translate into using a single dataset divided into two parts. To begin with, model development often involves a period of experimentation: trying out different models with different hyperparameters, and preprocessing the data in different ways. To avoid overfitting (see \nameref{validation}), this process requires a separate validation set, i.e.\ an additional set of training data that's not used directly in training or testing models. If you have no prior idea of what modelling approach you're going to use, then this experimentation phase could potentially involve a lot of comparisons. Due to the multiplicity effect (see \nameref{multcomparisons}), the more comparisons you do, the more likely you are to overfit the validation data, and so the less useful the validation set will become in guiding your modelling decisions. So, in practice you might want to set aside multiple validation sets for this. Then there's the question of how you adequately test your selected model. Because it has the same biases as the training data, a test set taken from the same dataset as the training data may not be sufficient to measure the model's generality --- see \nameref{testsets} and \nameref{measure} for more on this --- meaning that, in practice, you may need more than one test dataset to robustly evaluate your model. Also be aware that you will often need additional test data when using cross-validation; see \nameref{savedata}.


\subsection{Do take the time to understand your data}
Eventually you will want to publish your work. This is a lot easier to do if your data is from a reliable source, has been collected using a reliable methodology, and is of good quality. For instance, if you are using data collected from an internet resource, make sure you know where it came from. Is it described in a paper? If so, take a look at the paper; make sure it was published somewhere reputable, and check whether the authors mention any limitations of the data. Do not assume that, because a data set has been used by a number of papers, it is of good quality --- sometimes data is used just because it is easy to get hold of, and some widely used data sets are known to have significant limitations (see \cite{paullada2021data} for a discussion of this). If you train your model using bad data, then you will most likely generate a bad model: a process known as \textbf{garbage in garbage out}. One way to avoid bad data sets is to build a direct relationship with people who generate data, since this increases the likelihood of obtaining a good-quality dataset that meets your needs. It also avoids problems of overfitting community benchmarks; see \nameref{community}. Yet regardless of where your data comes from, always begin by making sure that your data makes sense. Do some \textbf{exploratory data analysis} (see \cite{cox2017exploratory} for suggestions). Look for missing or inconsistent records. It is much easier to do this now, before you train a model, rather than later, when you're trying to explain to reviewers why you used bad data.

\subsection{Don't look at \textit{all} your data}
As you look at data, it is quite likely that you will spot patterns and make insights that guide your modelling. This is another good reason to look at data. However, it is important that you do not make untestable assumptions that will later feed into your model. The ``untestable'' bit is important here; it's fine to make assumptions, but these should only feed into the training of the model, not the testing. So, to ensure this is the case, you should avoid looking closely at any test data in the initial exploratory analysis stage. Otherwise you might, consciously or unconsciously, make assumptions that limit the generality of your model in an untestable way. This is a theme I will return to several times, since the leakage of information from the test set into the training process is a common reason why ML models fail to generalise. See \nameref{leakage} for more on this.

\subsection{Do clean your data} \label{clean}
Even good-quality datasets will have issues. Some of these come from unavoidable noise or omissions in the data collection process, others are due to human error during collection or collation. Whatever the cause, it's important to identify any issues, and do this before you start to build models. One common problem to look out for is data duplication, i.e.\ the unintentional inclusion of multiple copies of a data point. This can cause serious problems when a model is evaluated (see \nameref{badaugmentation} for an example), so should be identified and removed early on. Another common problem is missing values. Some models can cope with these, but many can't, and so you'll have to replace missing values with something else before they can be trained. There are various forms of \textbf{imputation} that can be used to achieve this; see \cite{emmanuel2021survey} for a review. If you do imputation, be careful to avoid data leaks during imputation --- see \nameref{leakage}. You should also check for outliers in your data, but only remove these if they are likely to be the result of noise or error rather than being natural extremes of the underlying data-generating process. For example, if a person's age is greater than 150, then it's probably an error; if it's 110, then it could be a natural outlier. A related issue is meaningless or inconsistent data, for instance a person with a negative age. Data cleaning can be a time-consuming process, and becomes more challenging as the complexity of data increases. For this reason, many people have explored automating data cleaning using ML approaches; see \cite{cote2024data} for a review.

\subsection{Do make sure you have enough data} \label{augmentation}
If you don't have enough data, then it may not be possible to train a model that generalises. Working out whether this is the case can be challenging, and may not be evident until you start building models: it all depends on the signal to noise ratio in the data set. If the signal is strong, then you can get away with less data; if it's weak, then you need more data. If you can't get more data --- and this is a common issue in many research fields --- then you can try using \textbf{data augmentation} techniques (see \cite{wang2024comprehensive}, and for time series data, \cite{iglesias2023data}). These can be quite effective for boosting small data sets, though \nameref{badaugmentation}. Data augmentation is also useful in situations where you have  limited data in certain parts of your data set, e.g.\ in classification problems where you have less samples in some classes than others, a situation known as \textbf{class imbalance}. See \cite{haixiang2017learning} for a review of methods for dealing with this; also see \nameref{accuracy}. Another option for dealing with small data sets is to use transfer learning --- see \nameref{trends}. A danger when using small datasets is that different data partitions may be biased, for instance in terms of the quality or difficulty of data they contain. For this reason, it is advisable to consider frequent repartitioning. Cross-validation (see \nameref{multiple}) is an efficient way of achieving this in small data sets. If you have limited data, then it's also likely that you will have to limit the complexity of the ML models you use, since models with many parameters, like deep neural networks, can easily overfit small data sets (see \nameref{dnn}). Regardless of how you handle the problem of limited data, it's important to identify this issue early on, and come up with a suitable strategy to mitigate against it.

\subsection{Do talk to domain experts}
Domain experts can be very valuable. They can help you to understand which problems are useful to solve, they can help you choose the most appropriate feature set and ML model to use, and they can help you publish to the most appropriate audience. Failing to consider the opinion of domain experts can lead to projects which don't solve useful problems, or which solve useful problems in inappropriate ways. An example of the latter is using an opaque ML model to solve a problem where there is a strong need to understand how the model reaches an outcome, e.g.\ in making medical or financial decisions (see \cite{rudin2019stop}). At the beginning of a project, domain experts can help you to understand the data, and point you towards features that are likely to be predictive. At the end of a project, they can help you to publish in domain-specific journals, and hence reach an audience that is most likely to benefit from your research.

\subsection{Do survey the literature}
You're probably not the first person to throw ML at a particular problem domain, so it's important to understand what has and hasn't  been done previously. Other people having worked on the same problem isn't a bad thing; academic progress is typically an iterative process, with each study providing information that can guide the next. It may be discouraging to find that someone has already explored your great idea, but they most likely left plenty of avenues of investigation still open, and their previous work can be used as justification for your work. To ignore previous studies is to potentially miss out on valuable information. For example, someone may have tried your proposed approach before and found fundamental reasons why it won't work (and therefore saved you a few years of frustration), or they may have partially solved the problem in a way that you can build on. So, it's important to do a literature review before you start work; leaving it too late may mean that you are left scrambling to explain why you are covering the same ground or not building on existing knowledge when you come to write a paper.

\subsection{Do think about how your model will be deployed} \label{deployment}
Why do you want to build an ML model? This is an important question, and the answer should influence the process you use to develop your model. Many academic studies are just that --- studies --- and not really intended to produce models that will be used in the real world. This is fair enough, since the process of building and analysing models can itself give very useful insights into a problem. However, for many academic studies, the eventual goal is to produce an ML model that can be deployed in a real world situation. If this is the case, then it's worth thinking early on about how it is going to be deployed. For instance, if it's going to be deployed in a resource-limited environment, such as a sensor or a robot, this may place limitations on the complexity of the model. If there are time constraints, e.g.\ a classification of a signal is required within milliseconds, then this also needs to be taken into account when selecting a model. If using deep learning, then energy costs and carbon footprint may be a consideration, and if using LLMs, there may be further operational costs for hosting or accessing foundation models. Another consideration is how the model is going to be tied into the broader software system within which it is deployed; this procedure is often far from simple (see \cite{sculley2015hidden}). However, emerging approaches such as  \textbf{ML Ops} aim to address some of the difficulties; see \cite{kreuzberger2023machine} for a review, and \cite{shankar2022operationalizing} for a discussion of common challenges when operationalising ML models.

\section{How to reliably build models} \label{build}

Building models is one of the more enjoyable parts of ML. With modern ML frameworks, it's easy to throw all manner of approaches at your data and see what sticks. However, this can lead to a disorganised mess of experiments that's hard to justify and hard to write up. So, it's important to approach model building in an organised manner, making sure you use data correctly, and putting adequate consideration into the choice of models.

\subsection{Don't allow test data to leak into the training process} \label{leakage}

\begin{figure}[tb]
	\centering
	\includegraphics[width=0.46\columnwidth]{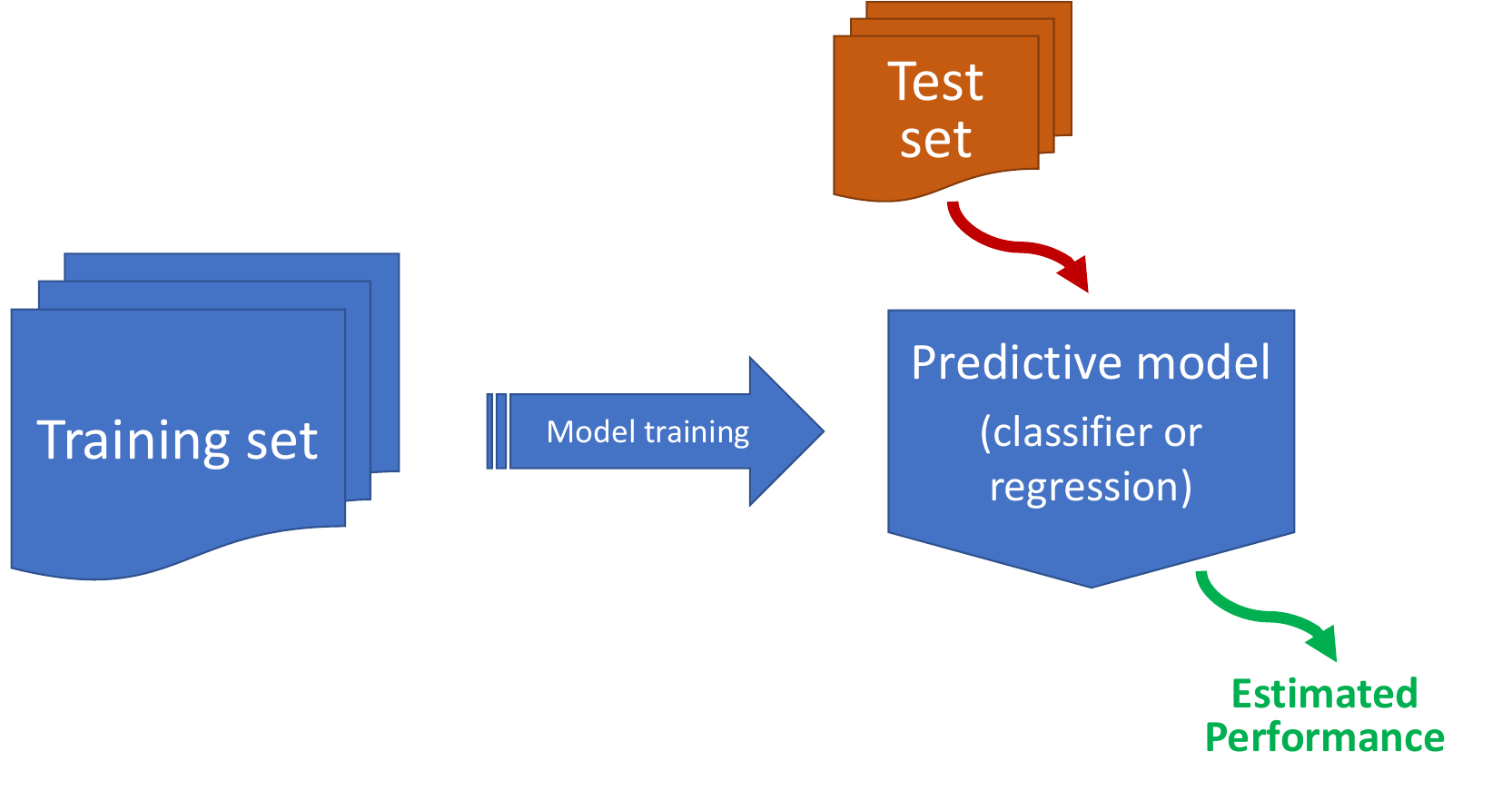}\hspace{0.04\columnwidth}
	\includegraphics[width=0.46\columnwidth]{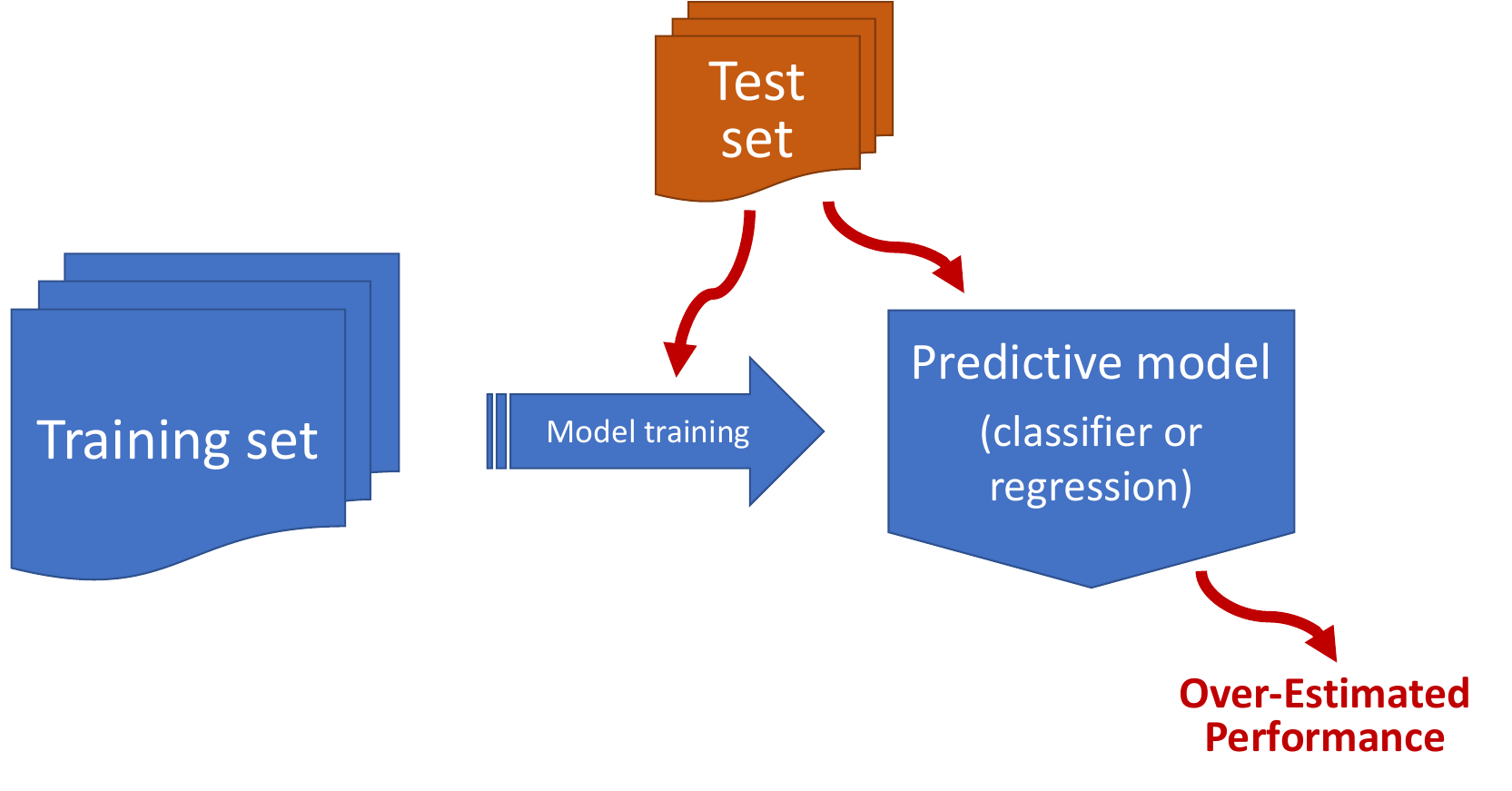}
	\caption{See \nameref{leakage}. [left] How things should be, with the training set used to train the model, and the test set used to measure its generality. [right] When there's a data leak, the test set can implicitly become part of the training process, meaning that it no longer provides a reliable measure of generality.}
	\label{fig:mlprocess}
\end{figure}

It's essential to have data that you can use to measure how well your model generalises. A common problem is allowing information about this data to leak into the configuration, training or selection of models (see Figure \ref{fig:mlprocess}). When this happens, the data no longer provides a reliable measure of generality, and this is a common reason why published ML models often fail to generalise to real world data. There are a number of ways that information can leak from a test set. Some of these seem quite innocuous. For instance, during data preparation, using information about the means and ranges of variables within the whole data set to carry out variable scaling or imputation --- in order to prevent information leakage, these statistics should be calculated using only the training data. Other common examples of information leakage are carrying out feature selection before partitioning the data (see \nameref{feature}), using the same test data to evaluate the generality of multiple models (see \nameref{validation} and \nameref{community}), and applying data augmentation before splitting off the test data (see \nameref{badaugmentation}). The best thing you can do to prevent these issues is to partition off a subset of your data right at the start of your project, and only use this independent test set once to measure the generality of a single model at the end of the project (see \nameref{savedata}). There are also forms of data leakage which are specific to certain types of data. Time series data is particularly problematic, since the order of samples is significant, and random splits can easily cause leakage and overfitting --- see \nameref{temporal} for more on this. Even for non-time series data, the experimental conditions used to generate data sets may lead to temporal dependencies, or other problematic conditions such as duplicated or similar samples --- see \nameref{testsets} for an example. In order to prevent leakage, these kinds of issues need to be identified and taken into account when splitting data. For a broader discussion of data leakage, see \cite{kapoor2023leakage}.

\subsection{Do try out a range of different models} \label{range}
Generally speaking, there's no such thing as a single best ML model. In fact, there's a proof of this, in the form of the No Free Lunch theorem, which shows that no ML approach is any better than any other when considered over every possible problem \citep{wolpert2002supervised}. So, your job is to find the ML model that works well for your particular problem. There is some guidance on this. For example, you can consider the \textbf{inductive biases} of ML models; that is, the kind of relationships they are capable of modelling. For instance, linear models, such as linear regression and logistic regression, are a good choice if you know there are no important non-linear relationships between the features in your data, but a bad choice otherwise. Good quality research on closely related problems may also be able to point you towards models that work particularly well. However, a lot of the time you're still left with quite a few choices, and the only way to work out which model is best is to try them all. Fortunately, modern ML libraries, such as scikit-learn \citep{varoquaux2015scikit} in Python, tidymodels \citep{kuhn2020tidymodels} in R, and MLJ \citep{blaom2020mlj} in Julia, allow you to try out multiple models with only small changes to your code, so there's no reason not to try them all out and find out for yourself which one works best. However, \nameref{inappropriate}, and use a validation set, rather than the test set, to evaluate them (see \nameref{validation}). When comparing models, \nameref{hyperparameters} and \nameref{multiple} to make sure you're giving them all a fair chance, and \nameref{multcomparisons} when you publish your results.

\subsection{Don't use inappropriate models} \label{inappropriate}
By lowering the barrier to implementation, modern ML libraries also make it easy to apply inappropriate models to your data. This, in turn, could look bad when you try to publish your results. A simple example of this is applying models that expect categorical features to a dataset containing numerical features, or vice versa. Some ML libraries allow you to do this, but it may result in a poor model due to loss of information. If you really want to use such a model, then you should transform the features first; there are various ways of doing this, ranging from simple one-hot encodings to complex learned embeddings. Other examples of inappropriate model choice include using a classification model where a regression model would make more sense (or vice versa), attempting to apply a model that assumes no dependencies between variables to time series data, or using a model that is unnecessarily complex (see \nameref{dnn}). Also, if you're planning to use your model in practice, \nameref{deployment}, and don't use models that aren't appropriate for your use case.

\subsection{Do keep up with progress in deep learning (and its pitfalls)}\label{trends}

\begin{figure}[tb!]
	\centering
	\includegraphics[width=\columnwidth]{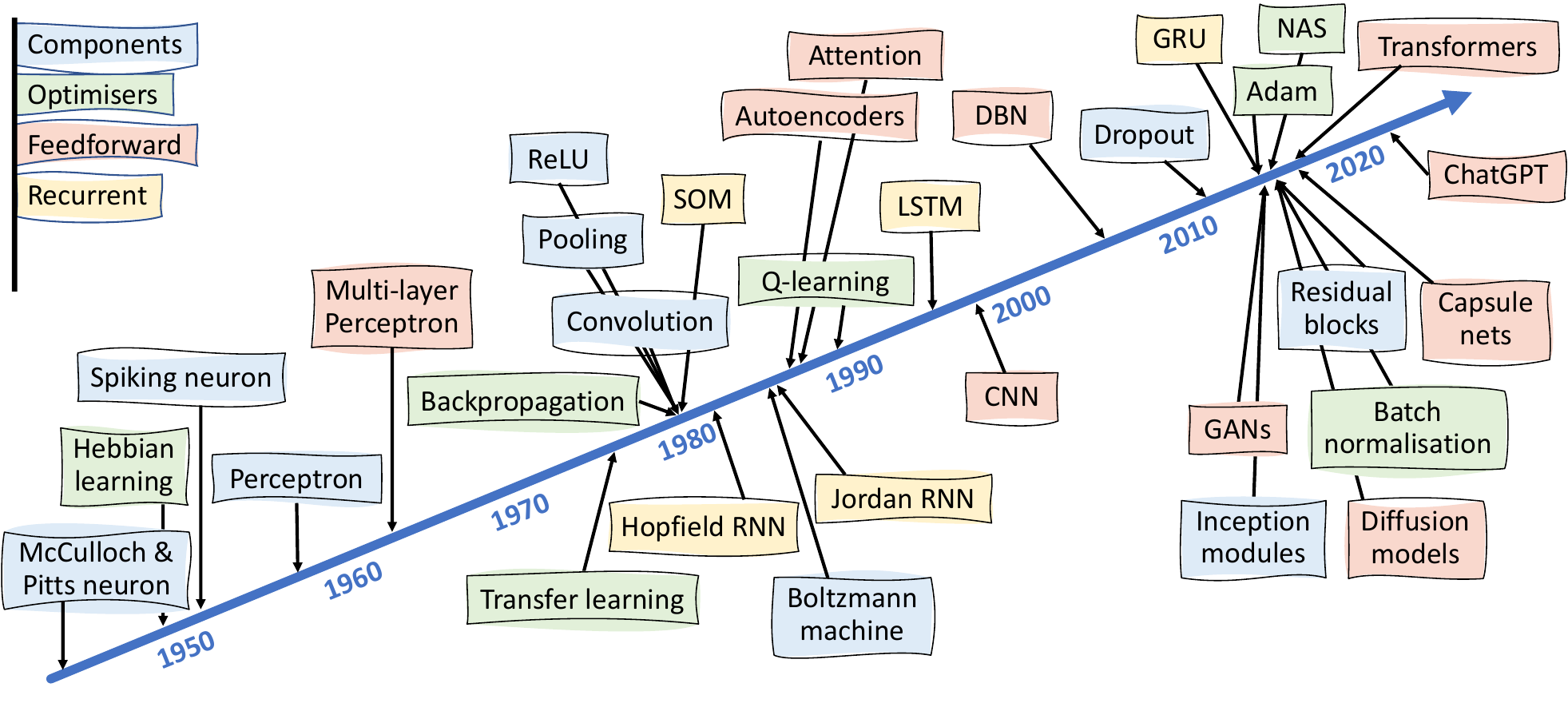}
	\caption{See \nameref{trends}. A rough history of neural networks and deep learning, showing what I consider to be the milestones in their development. For a far more thorough account of the field's historical development, take a look at \cite{schmidhuber2015deep}.}
	\label{fig:deeplearningtimeline}
\end{figure}

Whilst deep learning may not always be the best solution (see \nameref{dnn}), if you are going to use deep learning, then it's advisable to try and keep up with recent developments in this fast-moving field. Figure \ref{fig:deeplearningtimeline} summarises some of the important developments over time. Multilayer perceptrons (MLP) and recurrent neural networks (particularly LSTM) have been around for some time, but have largely been subsumed by newer models such as convolutional neural networks (CNN) \citep{li2021survey} and \textbf{transformers} \citep{lin2022survey}. For example, transformers have become the go-to model for processing sequential data (e.g.\ natural language), and are increasingly being applied to other data types too, such as images \citep{khan2022transformers}. A prominent downside of both transformers and deep CNNs is that they have many parameters and therefore require a lot of data to train them. However, an option for small data sets is to use \textbf{transfer learning}, where a model is pre-trained on a large generic data set and then fine-tuned on the data set of interest \citep{han2021pre}. Larger pre-trained models, many of which are freely shared on websites such as \href{https://huggingface.co}{Hugging Face}, are known as \textbf{foundation models}; see \cite{zhou2023comprehensive} for a survey. Whilst powerful, these come with their own set of pitfalls. For example, their ability to fully memorise input data is the cause of data security and privacy concerns \citep{li2023privacy}. The use of opaque, often poorly documented, training datasets also leads to pitfalls when fitting them into broader ML pipelines (see \nameref{ensemble} for more info) and comparing them fairly with other ML models (see \nameref{fair} and \nameref{community}). For an extensive, yet accessible, guide to deep learning, see \cite{zhang2023dive}.

\subsection{Don't assume deep learning will be the best approach}\label{dnn}
A common pitfall is to assume that deep neural networks will provide the best solution to any problem, and consequently fail to try out other, possibly more appropriate, models. Whilst deep learning is great for certain tasks, it is not good at everything; there are plenty of examples of it being out-performed by ``old fashioned'' machine learning models such as random forests and SVMs. See, for instance, \cite{grinsztajn2022tree}, who show that tree-based models often outperform deep learners on tabular data. Certain kinds of deep neural network architecture may also be ill-suited to certain kinds of data: see, for example, \cite{zeng2023transformers}, who argue that transformers are not well-suited to time series forecasting. There are also theoretical reasons why any one kind of model won't always be the best choice (see \nameref{range}). In particular, a deep neural network is unlikely to be a good choice if you have limited data, if domain knowledge suggests that the underlying pattern is quite simple, or if the model needs to be interpretable. This last point is particularly worth considering: a deep neural network is essentially a very complex piece of decision making that emerges from interactions between a large number of non-linear functions. Non-linear functions are hard to follow at the best of times, but when you start joining them together, their behaviour gets very complicated very fast. Whilst explainable AI methods (see \nameref{look}) can shine some light on the workings of deep neural networks, they can also mislead you by ironing out the true complexities of the decision space (see \cite{molnar2020general}). For this reason, you should take care when using either deep learning or explainable AI for models that are going to make high stakes or safety critical decisions; see \cite{rudin2019stop} for more on this.

\subsection{Do be careful where and how you do feature selection} \label{feature}

\begin{figure}[!ptb]
	\centering
	\includegraphics[width=0.8\columnwidth]{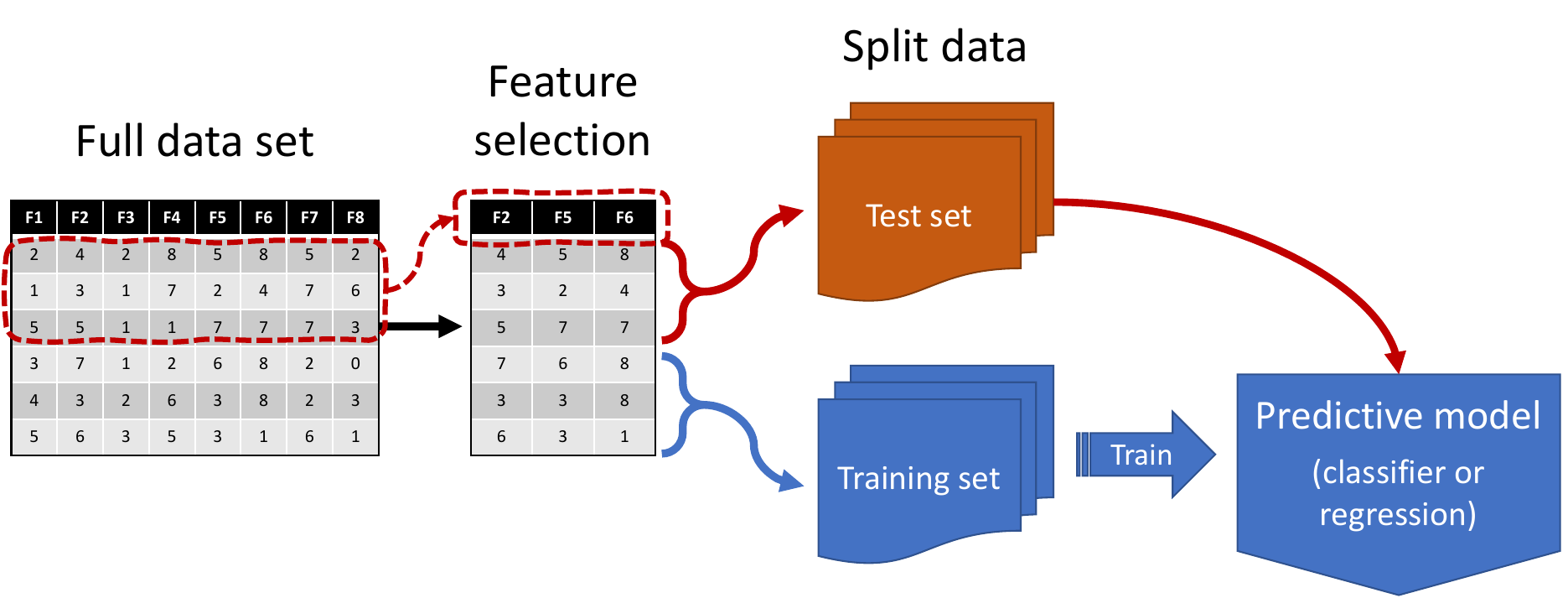}\\\vspace{1.15cm}
	\includegraphics[width=0.8\columnwidth]{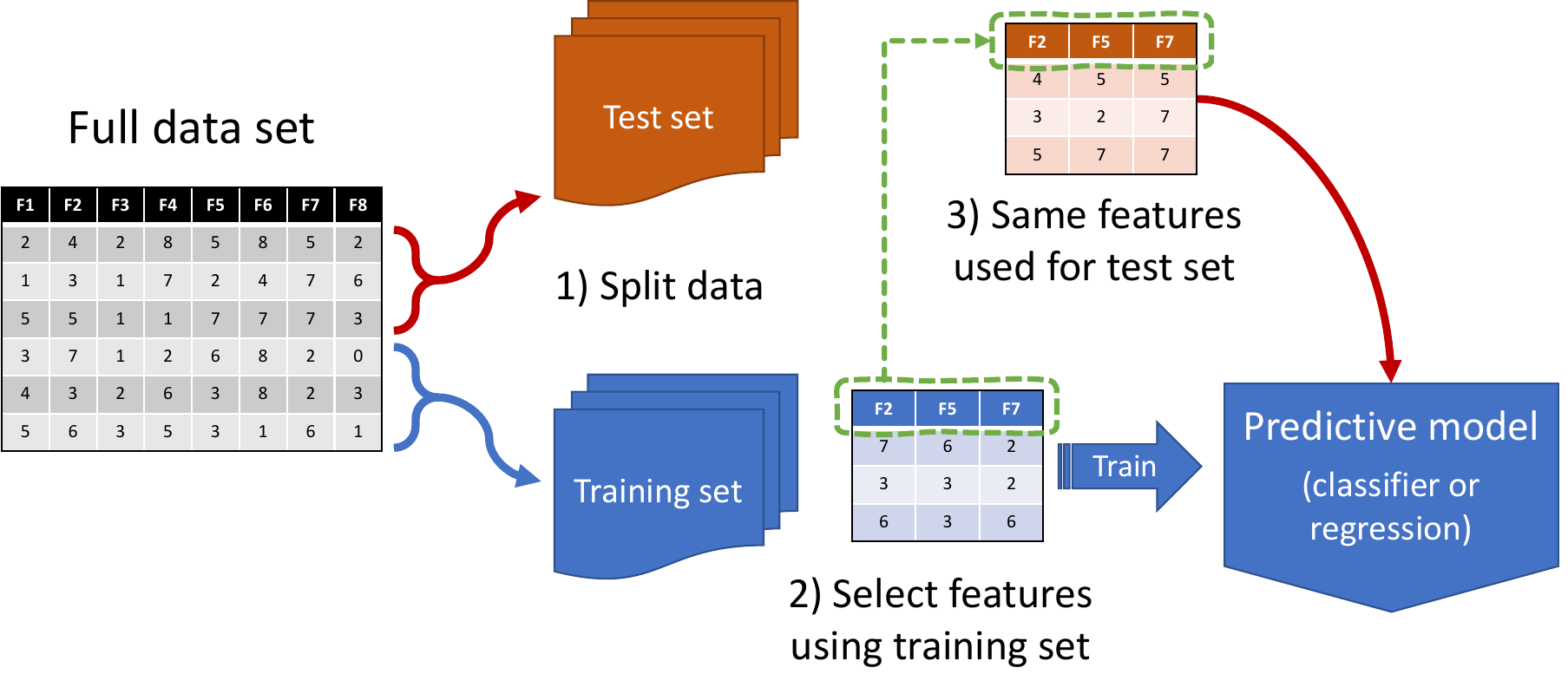}\\\vspace{1.15cm}
	\includegraphics[width=0.95\columnwidth]{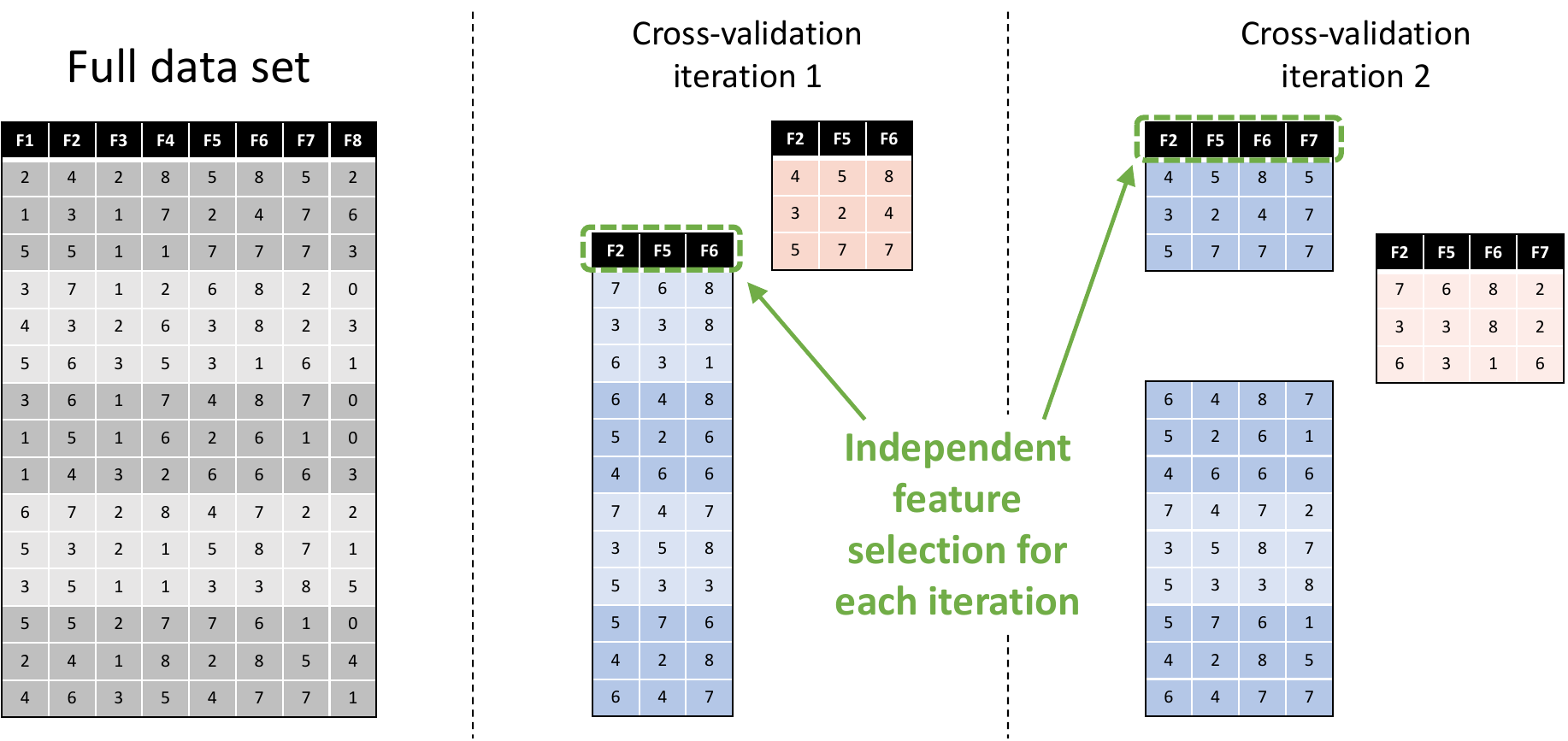}\vspace{5mm}
	\caption{See \nameref{feature}. [top] Data leakage due to carrying out feature selection before splitting off the test data (outlined in red), causing the test set to become an implicit part of model training. [centre] How it should be done. [bottom] When using cross-validation, it's important to carry out feature selection independently for each iteration, based only on the subset of data (shown in blue) used for training during that iteration.}
	\label{fig:featureselection}
\end{figure}

A common stage of training a model is to carry out \textbf{feature selection} (surveyed by \cite{cai2018feature}). When doing this, it is important to treat it as part of model training, and not something more general that you do before model training. A particularly common error is to do feature selection on the whole data set before splitting off the test set, something that will result in information leaking from the test set into the training process (see \nameref{leakage}). Instead, you should only use the training set to select the features which are used in both the training set and the test set (see Figure \ref{fig:featureselection}). The same is true when doing \textbf{dimensionality reduction}. For example, if you're using principal component analysis (PCA), the component weightings should be determined by looking only at the training data; the same weightings should then be applied to the test set. Special care should be taken when using autoencoders for dimensionality reduction --- see \nameref{ensemble}. If you're doing cross-validation (see \nameref{multiple}) then it's important to carry out feature selection or dimensionality reduction independently within each iteration, each time using just the training folds (see Figure \ref{fig:featureselection}, bottom).

\subsection{Do optimise your model's hyperparameters} \label{hyperparameters}
Many models have \textbf{hyperparameters} --- that is, numbers or settings that affect the configuration of the model. Examples include the kernel function used in an SVM, the number of trees in a random forest, and the architecture of a neural network. Many of these hyperparameters significantly effect the performance of the model, and there is generally no one-size-fits-all. That is, they need to be fitted to your particular data set in order to get the most out of the model. Whilst it may be tempting to fiddle around with hyperparameters until you find something that works, this is not likely to be an optimal approach. It's much better to use some kind of \textbf{hyperparameter optimisation} strategy, and this is much easier to justify when you write it up. Basic strategies include random search and grid search, but these don't scale well to large numbers of hyperparameters or to models that are expensive to train, so it's worth using tools that search for optimal configurations in a more intelligent manner. See \cite{bischl2023hyperparameter} for further guidance. It is also possible to use \textbf{AutoML} techniques to optimise both the choice of model and its hyperparameters, in addition to other parts of the machine learning pipeline --- see \cite{barbudo2023eight} for a review. 

\subsection{Do avoid learning spurious correlations} \label{spurious}

\begin{figure}[!tb]
	\centering
	\includegraphics[width=0.85\columnwidth]{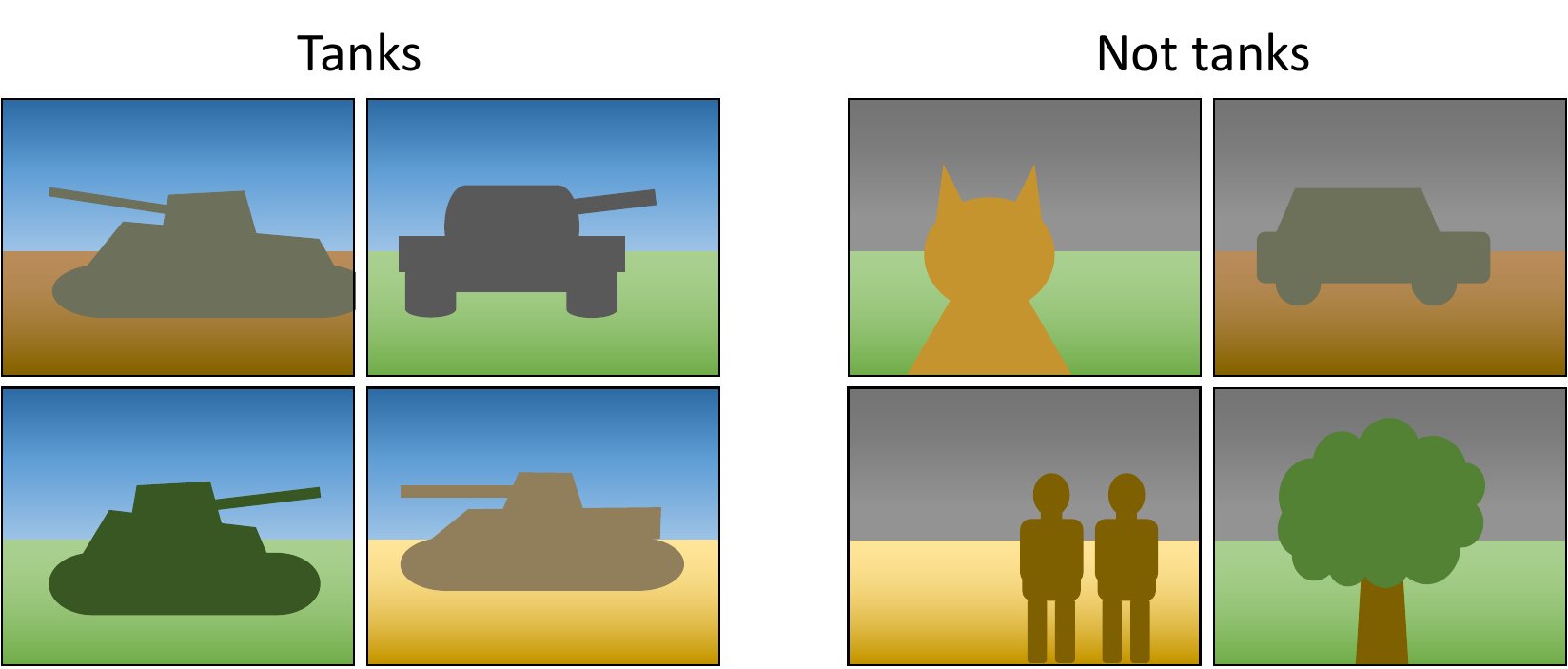}
	\caption{See \textbf{\nameref{spurious}}. The problem of spurious correlations in images, as illustrated by the tank problem. The images on the left are tanks, and those on the right are not tanks. However, the consistent background (blue for tanks, grey for others) means that these images can be classified by merely looking at the colours of pixels towards the top of the images, rather than having to recognise the objects in the images, resulting in a poor model.}
	\label{fig:tanks}
\end{figure}

Spurious correlations are features within data which are correlated with the target variable, but which have no semantic meaning. They are basically red herrings, and it's not uncommon for ML models to pick up on them in training, and consequently fail to generalise well. A classic example is the tank problem. Legend\footnote{There is some debate about whether this actually happened: see \url{https://www.gwern.net/Tanks}.} has it that the US military were looking to train an ML model that could recognise tanks. However, because the tank pictures used in training were taken during different weather conditions to the non-tank pictures, the model ended up discriminating based on features such as the number of blue pixels in the sky, rather than the presence of a tank (see Figure \ref{fig:tanks} for an illustration). An ML model that uses such spurious correlations to perform classification would appear to be very good, in terms of its metric scores, but would not work in practice. More complex data tends to contain more of these spurious correlations, and more complex models have more capacity to overfit spurious correlations. This means that spurious correlations are a particular issue for deep learning, where approaches such as regularisation (see \nameref{trends}) and data augmentation (see \nameref{augmentation}) can help mitigate against this. However, spurious correlations can occur in all data sets and models, so it is always worth looking at your trained model to see whether it's responding to appropriate features within your data --- see \nameref{look}. 

\section{How to robustly evaluate models} \label{evaluate}

In order to contribute to progress in your field, you need to have valid results that you can draw reliable conclusions from. Unfortunately it's really easy to evaluate ML models unfairly, and, by doing so, muddy the waters of academic progress. So, think carefully about how you are going to use data in your experiments, how you are going to measure the true performance of your models, and how you are going to report this performance in a meaningful and informative way.

\subsection{Do use an appropriate test set}\label{testsets}
First of all, always use a test set to measure the generality of an ML model. How well a model performs on the training set is almost meaningless, and a sufficiently complex model can entirely learn a training set yet capture no generalisable knowledge. It's also important to make sure the data in the test set is appropriate. That is, it should not overlap with the training set and it should be representative of the wider population. For example, consider a photographic data set of objects where the images in the training and test set were collected outdoors on a sunny day. The presence of the same weather conditions means that the test set will not be independent, and by not capturing a broader variety of weather conditions, it will also not be representative. Similar situations can occur when a single piece of equipment is used to collect both the training and test data; if the model overlearns characteristics of the equipment, it will likely not generalise to other pieces of equipment, and this will not be detectable by evaluating it on the test set. If using public datasets to test a model, be wary of \textbf{Frankenstein datasets}, which are assembled from other public datasets and risk overlap with training data. Also be careful when handling datasets that contain multiple data points for each subject; if using these, it's important to make sure that each subject's data points are kept together when splitting off the test set or when doing cross-validation. See \cite{roberts2021common} for a revealing account of how a number of these pitfalls led to the failure of the vast majority of Covid-19 detection models to generalise beyond their test sets.

\subsection{Don't do data augmentation \textit{before} splitting your data}\label{badaugmentation}
Data augmentation (see \nameref{augmentation}) can be a useful technique for balancing datasets and boosting the generality and robustness of ML models. However, it's important to do data augmentation only on the training set, and not on data that's going to be used for testing. Including augmented data in the test set can lead to a number of problems. One problem is that the model may overfit the characteristics of the augmented data, rather than the original samples, and you won't be able to detect this if your test set also contains augmented data. A more critical problem occurs when data augmentation is applied to the entire data set before it is split into training and test sets. In this scenario, augmented versions of training samples may end up in the test set, which in the worst case can lead to a particularly nefarious form of data leakage in which the test samples are mostly variants of the training samples. For an interesting study of how this problem affected an entire field of research, see \cite{vandewiele2021overly}.

\subsection{Do avoid sequential overfitting} \label{validation}

\begin{figure}[!tb]
	\centering
	\includegraphics[width=0.9\columnwidth]{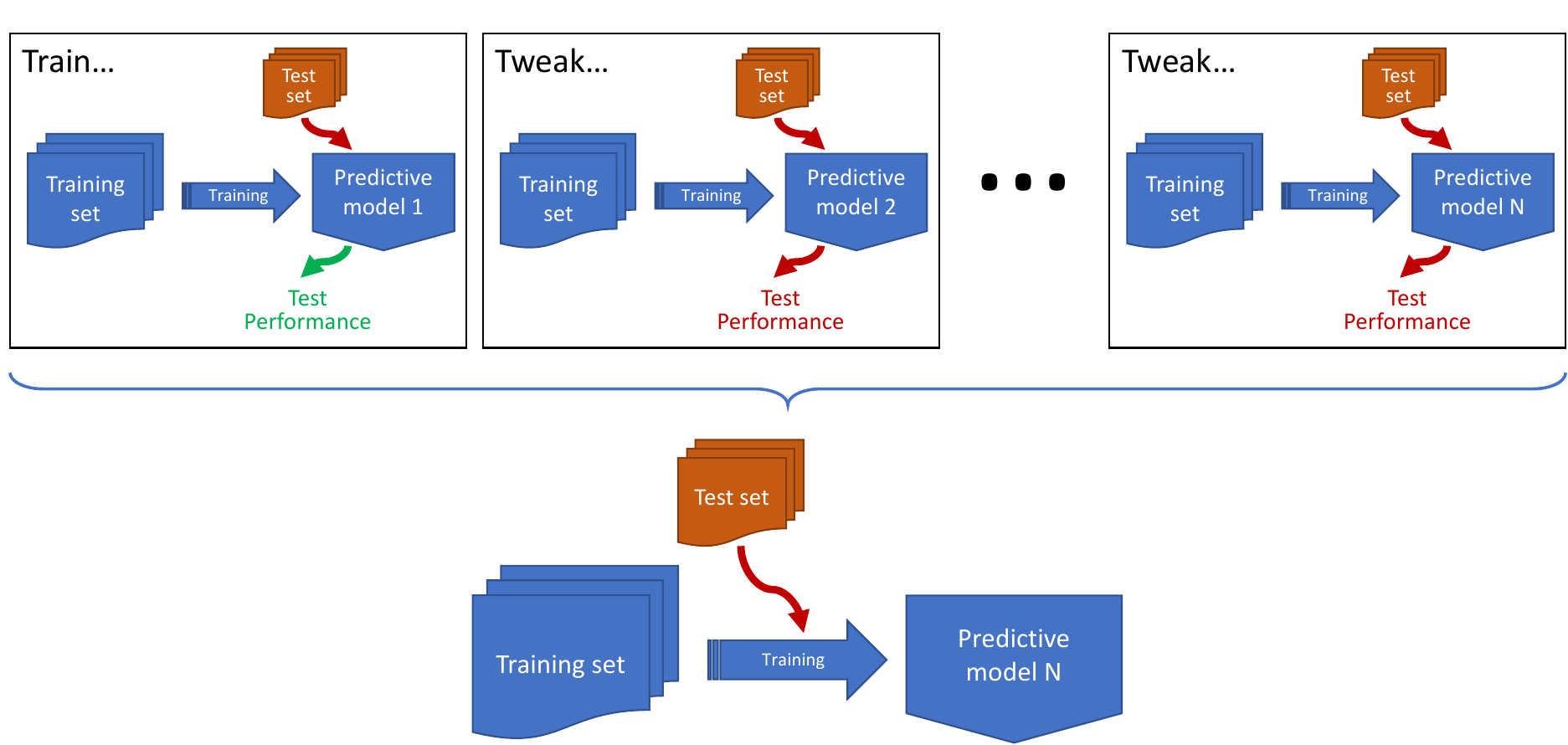}\\\vspace{1cm}
	\includegraphics[width=0.9\columnwidth]{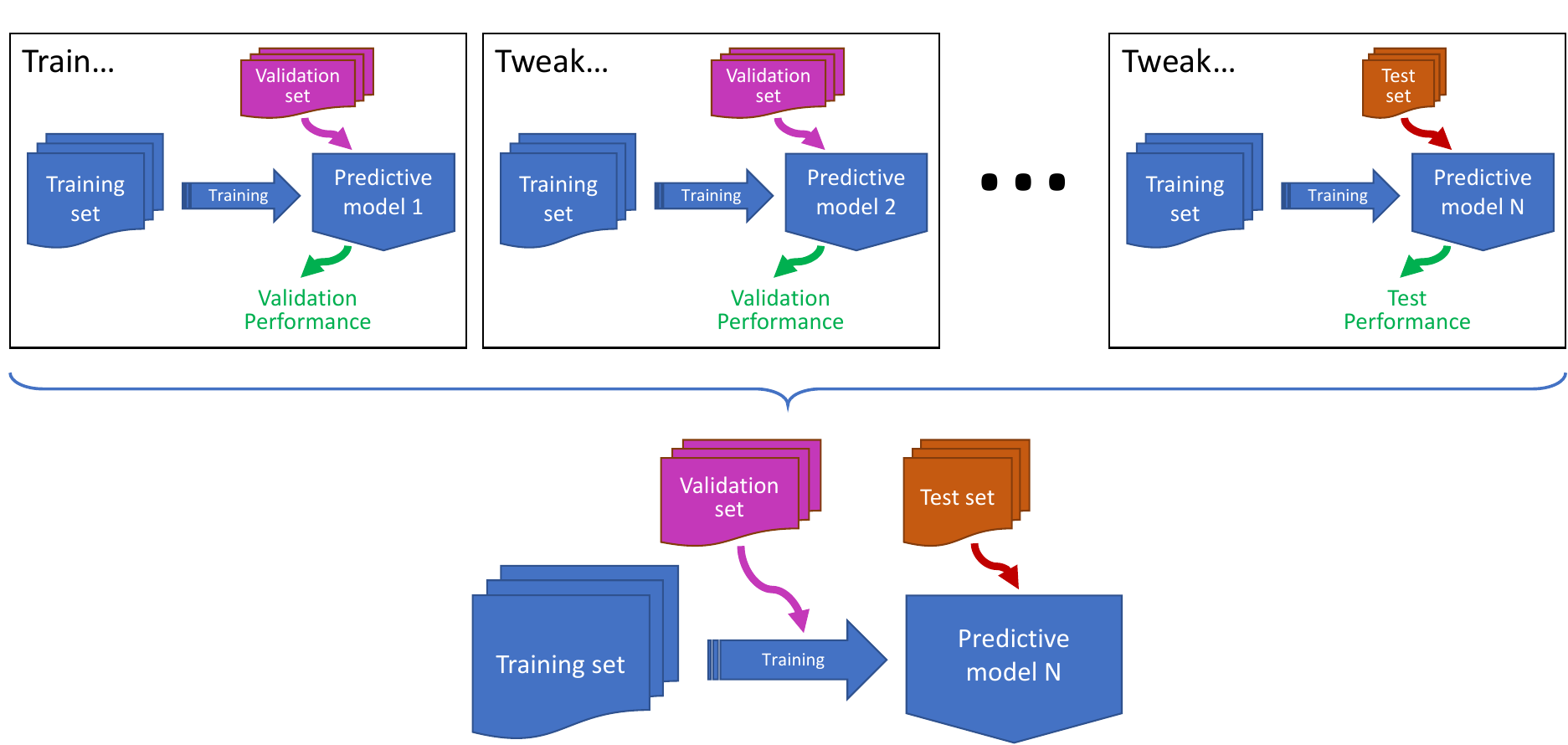}
	\caption{See \nameref{validation}. [top] Using the test set repeatedly during model selection results in the test set becoming an implicit part of the training process. [bottom] A validation set should be used instead during model selection, and the test set should only be used once to measure the generality of the final model.}
	\label{fig:overfitting}
\end{figure}

Oddly, one of the most pernicious forms of data leakage doesn't have a commonly-agreed name\footnote{Though \cite{hosseini2020tried} suggested ``over-hyping", from \textit{over}fitting of \textit{hyp}er-parameters.}, so I’m going to refer to it as sequential overfitting. This occurs when you train multiple models in succession, using knowledge gained about each model's performance to guide the configuration of the next one, and you use the same test set to evaluate each model. Often this is done as an informal process, trying out different models and different hyperparameters until you get good performance on the test set. As such, it is rarely documented, which is one reason why it's so pernicious. Specifically, the problem lies in using the test set throughout this process, since using the test set to choose between models means that information about the test set implicitly leaks into the training process. See Figure \ref{fig:overfitting} for an illustration of this idea. The consequence is that models gradually overfit the test set; the more times you use the test set, the more the overfitting that occurs. The solution is to either use a \textbf{validation set} (i.e. a separate set of samples that are not directly used in training, but which are used to guide training) or use a holdout data set to test the final model. See \cite{cawley2010over} and \cite{hosseini2020tried} for more on this.

\subsection{Do evaluate a model multiple times} \label{multiple}
Many ML models are stochastic or unstable. That is, if you train them multiple times, or if you make small changes to the training data, then their performance varies significantly. The same is true of using LLMs at inference time. This means that a single evaluation of a model can be unreliable, and may either underestimate or overestimate the model's true potential. For this reason, it is common to carry out multiple evaluations. At training time, there are numerous ways of doing this. For stochastic models, the simplest is to train the same model multiple times using different random seeds and then look at the average performance. A more robust approach is to also vary the data for each model trained. \textbf{Cross-validation} (CV) is a particularly popular way of doing this, and comes in numerous flavours \citep{arlot2010survey}, most of which involve splitting the data into a number of folds. When doing CV, it is important to be aware of any dependencies within the data and take these into account. Failure to do so can result in data leakage. For instance, in medical datasets, it is commonplace to have multiple data points for a single subject; to avoid data leakage, these should be kept together within the same fold. Time series data is particularly problematic for CV; see {\nameref{temporal} for a discussion of how to handle this. If you're carrying out hyperparameter optimisation, then you should use \textbf{nested cross-validation} (also known as double cross-validation), which uses an extra loop inside the main cross-validation loop to avoid overfitting the test folds. If some of your data classes are small, then you may need to do \textbf{stratification}, which ensures each class is adequately represented in each fold. In addition to looking at average performance across multiple evaluations, it is also standard practice to provide some measure of spread or confidence, such as the standard deviation or the 95\% confidence interval.

\subsection{Do save some data to evaluate your final model instance} \label{savedata}
I've used the term \textit{model} quite loosely, but there is an important distinction between evaluating the potential of a general model (e.g.\ how well a neural network can solve your problem), and the performance of a particular model instance (e.g.\ a specific neural network produced by one run of back-propagation). Cross-validation is good at the former, but it's less useful for the latter.  Say, for instance, that you carried out ten-fold cross-validation. This would result in ten model instances. Say you then select the instance with the highest test fold score as the model which you will use in practice. How do you report its performance? Well, you might think that its test fold score is a reliable measure of its performance, but it probably isn't. First, the amount of data in a single fold is relatively small. Second, the instance with the highest score could well be the one with the easiest test fold, so the evaluation data it contains may not be representative. Consequently, the only way of getting a reliable estimate of the model instance's generality may be to use another test set. This is also true in situations where the independence of the existing test set may have been compromised, e.g.\ by using it more than once (see \nameref{validation}). So, if you have enough data, it's better to keep some aside and only use it once to provide an unbiased estimate of the final selected model instance. However, it's worth noting one other option when using cross-validation, and that is to ensemble the model instances (see \nameref{ensemble}). The resulting ensemble will have performance in line with the average as measured through cross-validation, so another test set is not required to measure its performance. On the downside, it will likely have poorer inference time, efficiency and interpretability than a single model instance, so this approach is generally only worth considering if you have very little data.

\subsection{Do choose metrics carefully} \label{accuracy}

\begin{figure}[!tb]
	\centering
	\includegraphics[width=0.9\columnwidth]{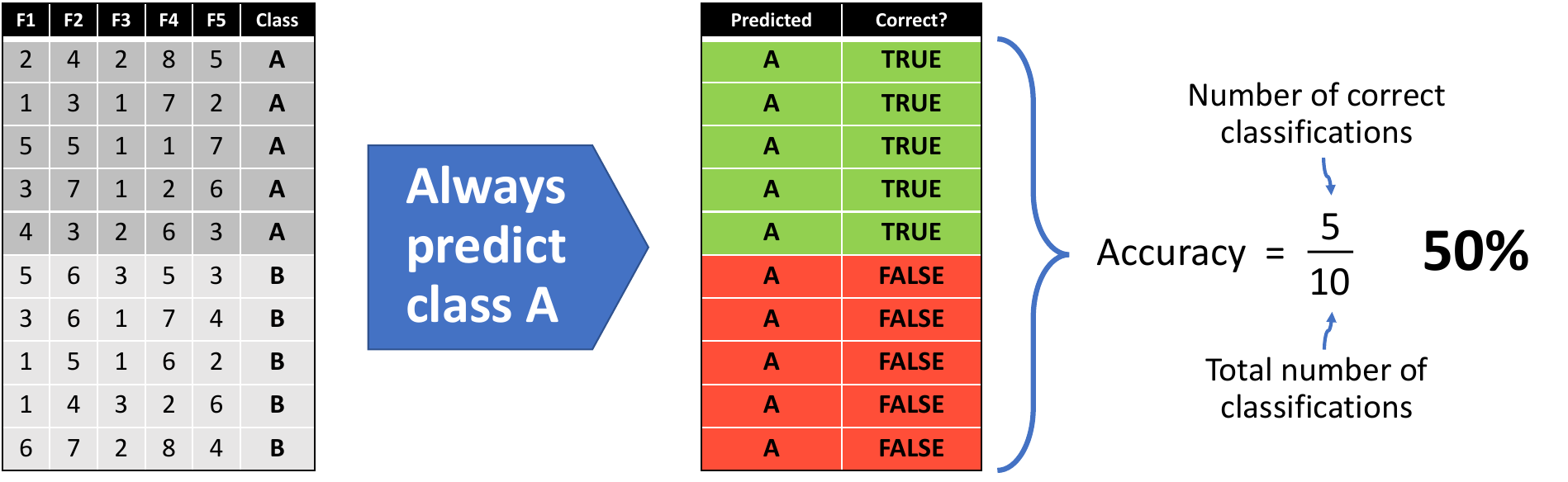}\\\vspace{0.5cm}
	\includegraphics[width=0.9\columnwidth]{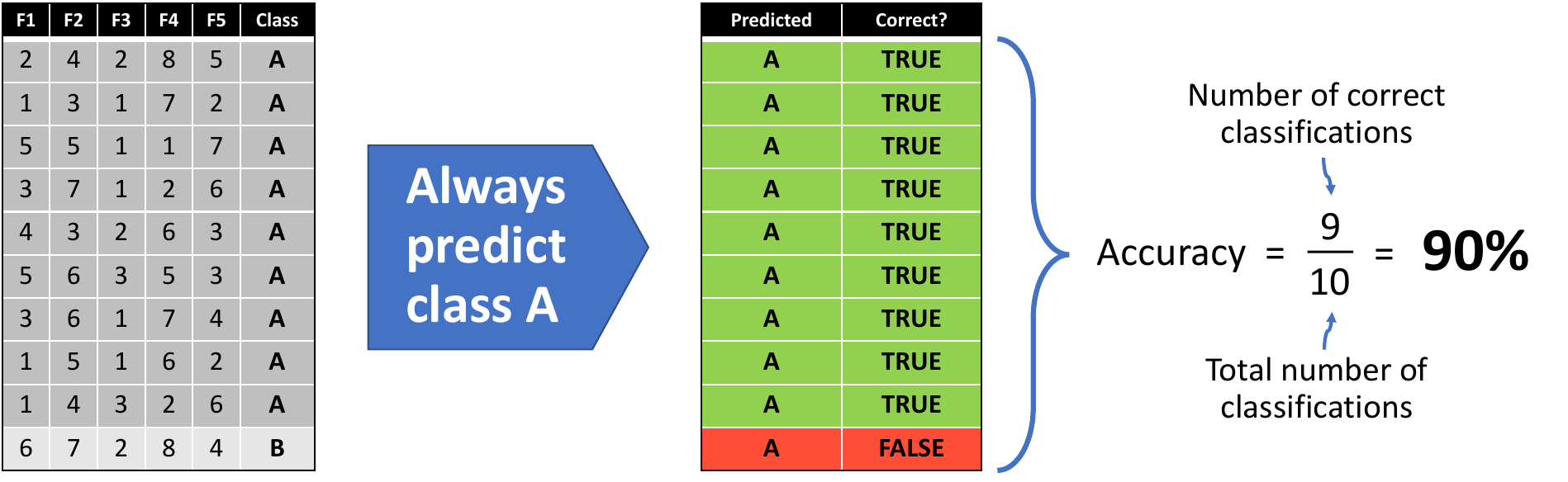}
	\caption{See \textbf{\nameref{accuracy}}. The problem with using accuracy as a performance metric on imbalanced data. Here, a dummy model which always predicts the same class label has an accuracy of 50\% or 90\% depending on the distribution of class labels within the data.}
	\label{fig:accuracy}
\end{figure}

Be careful which metrics you use to evaluate your ML models. For instance, in the case of classification models, the most commonly used metric is accuracy, which is the proportion of samples in the data set that were correctly classified by the model. This works fine if your classes are balanced, i.e.\ if each class is represented by a similar number of samples within the data set. But many data sets are not balanced, and in this case accuracy can be a very misleading metric. Consider, for example, a data set in which 90\% of the samples represent one class, and 10\% of the samples represent another class. A binary classifier which always outputs the first class, regardless of its input, would have an accuracy of 90\%, despite being completely useless (see Figure \ref{fig:accuracy}). In this kind of situation, it would be preferable to use a metric such as $F_1$ score, Cohen's kappa coefficient ($\kappa$) or Matthews Correlation Coefficient (MCC), all of which are relatively insensitive to class size imbalance. For a broader review of methods for dealing with imbalanced data, see \cite{haixiang2017learning}. There are also various pitfalls associated with regression metrics, particularly within the context of time series forecasting; see \cite{hewamalage2023forecast} for a discussion of these. A well-known example is relying only on the RMSE, which (a bit like accuracy) is susceptible to assigning high value to models which always predict no change. Also see \nameref{measure}.

\subsection{Do consider model fairness} \label{fairness}
Overall performance metrics are not the only important measures of how good a model is. If a model is to be deployed within the real world, another important measure is fairness. There are various definitions of fairness, but in a nutshell, it's about making sure that the model doesn't treat its human subjects unequally with regard to characteristics such as gender, ethnicity, income or personal politics. This is also referred to as \textbf{algorithmic bias}, and there are many examples of models being biased towards or against particular groups of people. A common source of unfairness is using an unrepresentative dataset to train an ML model. For instance, if a medical diagnosis model is trained on data from a single country, then the data may be biased towards the majority ethnicity, and the model may not operate fairly when exposed to users from other ethnicities. However, unfairness can also come from other sources, including subconscious bias during data preparation and the inductive biases of the model. Regardless of the source, it is important to understand any resulting biases, and ideally take steps to mitigate against them (e.g.\ applying data augmentation to minority samples --- see \nameref{augmentation}). There are many different fairness metrics, so part of the puzzle is working out which are most relevant to your modelling context; see \cite{caton2024fairness} for a review.

\subsection{Don't ignore temporal dependencies in time series data} \label{temporal}
Time series data is unlike many other kinds of data in that the order of the data points is important. Many of the pitfalls in handling time series data are a result of ignoring this fact. Most notably, time series data are subject to a particular kind of data leakage (see \nameref{leakage}) known as \textbf{look ahead bias}. This occurs when some or all of the data points used to train the model occur later in the time series than those used to test the model. In effect, this can allow knowledge of the future to leak into training, and this can then bias the test performance. A situation where this commonly occurs is when standard cross-validation (see \nameref{multiple}) is applied to time series data, since it results in the training folds in all but one of the cross-validation iterations containing data that is in the future relative to the test fold. This can be avoided by using special forms of cross-validation that respect temporal dependencies, such as \textbf{blocked cross-validation}, though whether this is necessary depends to some extent on the nature of the time series data, e.g. whether it is stationary or non-stationary. See \cite{cerqueira2020evaluating} and \cite{ruf2022information} for more on this. Look ahead bias can also result from carrying out data-dependent preprocessing operations before splitting off the test data; see Figure \ref{fig:scaling} for a simple example of this, but also see \nameref{feature}.

\begin{figure}[!tb]
	\centering
	\includegraphics[width=0.95\columnwidth]{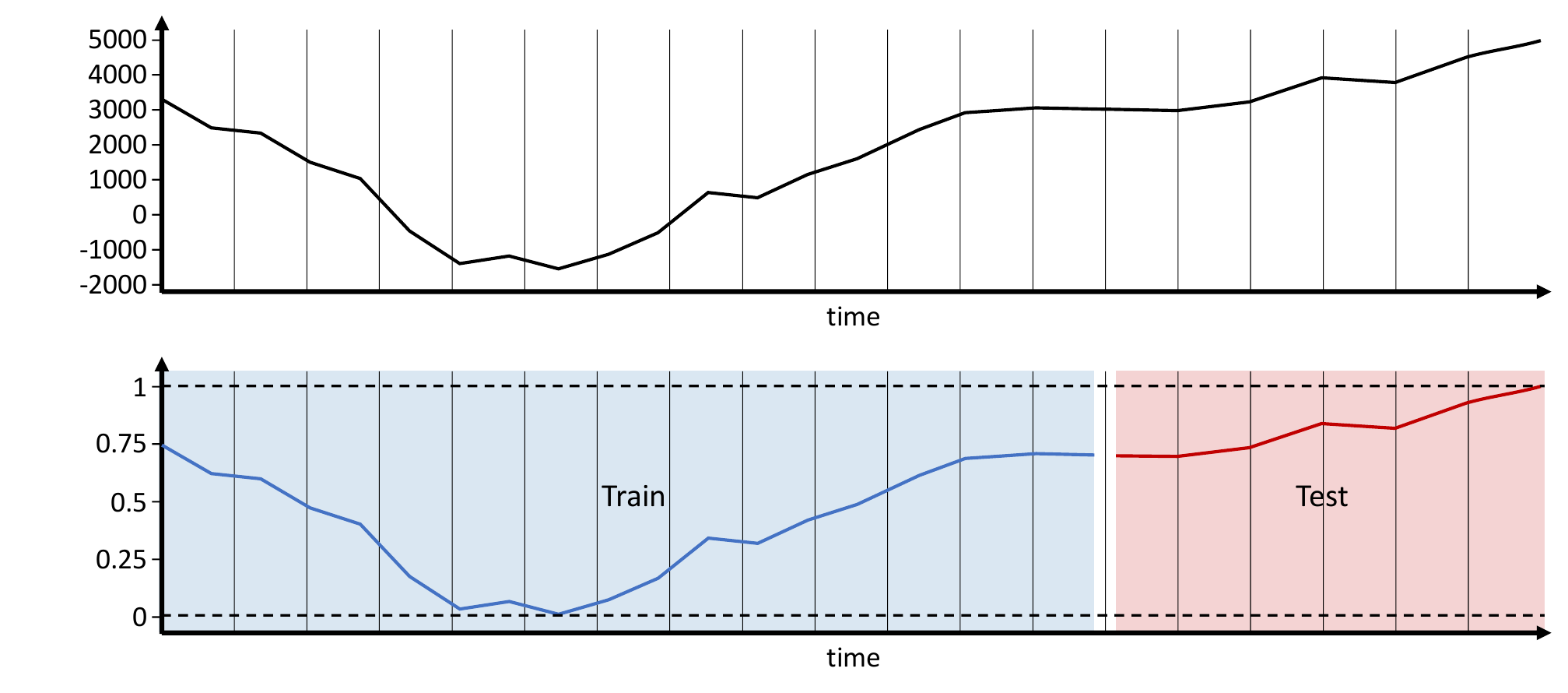}\\
	\vspace{1cm}
	\includegraphics[width=0.95\columnwidth]{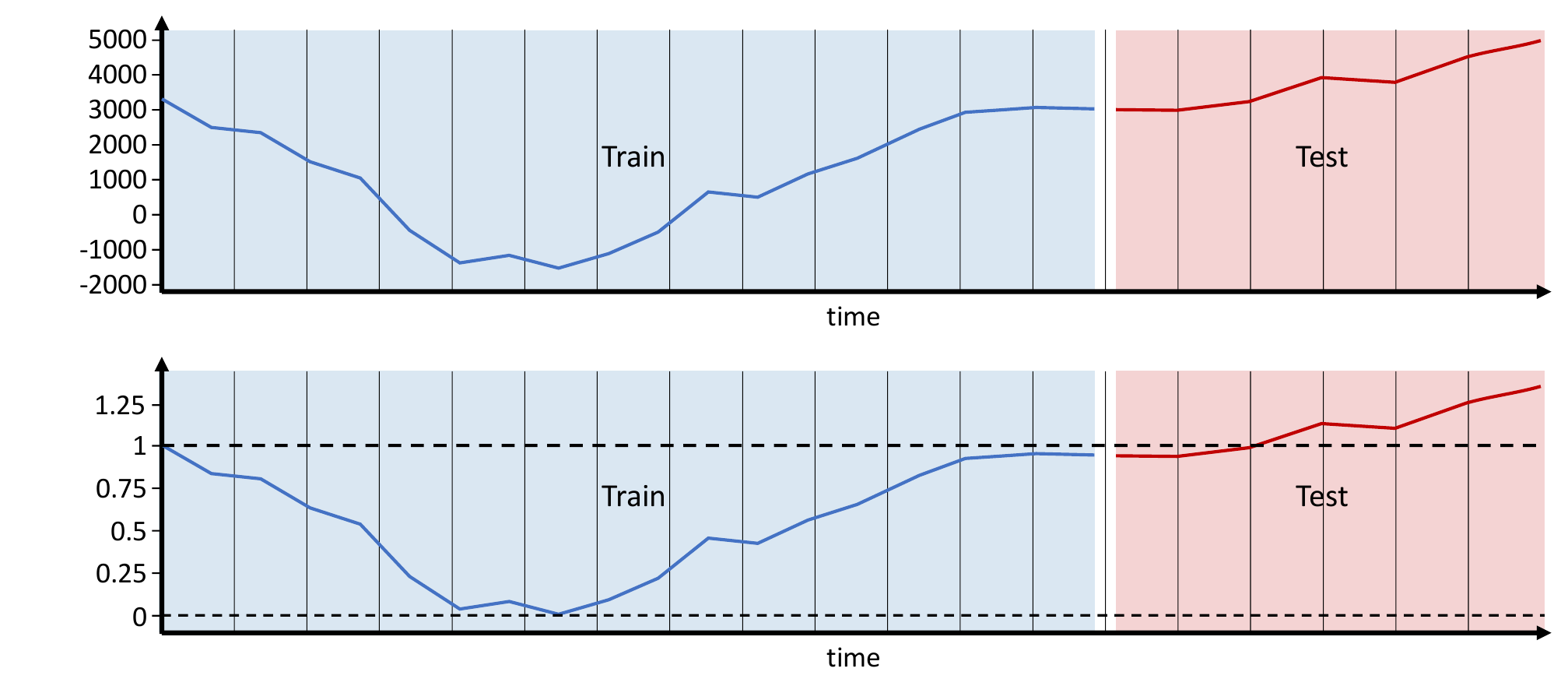}
	\caption{See \nameref{temporal}. [top] A time series is scaled to the interval $[0,1]$ before splitting off the test data (shown in red). This could allow the model to infer that values will increase in the future, causing a potential look ahead bias. [bottom] Instead, the data should be split before doing scaling, so that information about the range of the test data can't leak into the training data.}
	\label{fig:scaling}
\end{figure}

\section{How to compare models fairly} \label{compare}
Comparing models is the basis of academic research, but it's surprisingly difficult to get it right. If you carry out a comparison unfairly, and publish it, then other researchers may subsequently be led astray. So, do make sure that you evaluate different models within the same context, do explore multiple perspectives, and do use make correct use of statistical tests.

\subsection{Don't assume a bigger number means a better model} \label{fair}
It's not uncommon for a paper to state something like ``In previous research, accuracies of up to 94\% were reported. Our model achieved 95\%, and is therefore better.'' There are various reasons why a higher figure does not imply a better model. For instance, if the models were trained or evaluated on different partitions of the same data set, then small differences in performance may be due to this. If the datasets had different degrees of class imbalance, then the difference in accuracy could merely reflect this (see \nameref{accuracy}). If they used different data sets entirely, then this may account for even large differences in performance. Another reason for unfair comparisons is the failure to carry out the same amount of hyperparameter optimisation (see \nameref{hyperparameters}) when comparing models; for instance, if one model has default settings and the other has been optimised, then the comparison won't be fair. For these reasons, and others, comparisons based on published figures should always be treated with caution. To be sure of a fair comparison between two approaches, you should freshly implement all the models you're comparing, optimise each one to the same degree, carry out multiple evaluations (see \nameref{multiple}), and then use statistical tests (see \nameref{statistical}) to determine whether the differences in performance are significant. A further complication when comparing foundation models (see \nameref{trends}) it that the original training data is often unknown; consequently it may be impossible to ensure that the test set is independent of the training data, and therefore a fair basis for comparison.

\subsection{Do use meaningful baselines} \label{baselines}
When introducing a new modelling approach, it is essential to compare against established approaches. These are commonly referred to as \textbf{baseline models}, or just baselines. It is important that these baselines are selected so that they provide a meaningful basis for comparison. Baselines are often simpler than the new approach, and are chosen to demonstrate that any complexity in the new model is necessary. For example, if you're extending model X, then it makes sense to use model X as a baseline. However, it also makes sense to use other simpler models. For instance, if you're developing a deep learning approach that uses tabular data, then you should also compare against simpler models like decision trees and SVMs to show that a more complex approach is justified. If you're solving a regression problem, then you should also consider using simple baselines like logistic regression. The simplest baselines are known as \textbf{naïve baselines} and are used to show that your model is not doing something trivial. An illustrative example of why these are necessary is described in \cite{hewamalage2023forecast}, where a complex transformer model designed for time series forecasting is shown to perform worse than a naïve baseline that always forecasts the next value in a time series to be the same as the previous value. This kind of naïve baseline, in which there is no real decision making process, is also known as a \textbf{dummy model}. Another example is a classifier that always outputs the most frequent class label (as described in \nameref{accuracy}). In addition to simple baselines, it's also important to compare against state-of-the-art (\textbf{SOTA}) models. Otherwise you may be asked something like ``Why are you extending model X when model Y is known to be better than model X?"

\subsection{Do use statistical tests when comparing models} \label{statistical}
If you want to convince people that your model is better than someone else's, then a statistical test can be a useful tool. Broadly speaking, there are two categories of tests for comparing individual ML models. The first is used to compare individual model instances, e.g.\ two trained decision trees. For example, McNemar's test is a fairly common choice for comparing two classifiers, and works by comparing the classifiers' output labels for each sample in the test set (so do remember to record these). The second category of tests are used to compare two models more generally, e.g.\ whether a decision tree or a neural network is a better fit for the data. These require multiple evaluations of each model, which you can get by using cross-validation or repeated resampling (or, if your training algorithm is stochastic, multiple repeats using the same data). The test then compares the two resulting distributions. Student's T test is a common choice for this kind of comparison, but it's only reliable when the distributions are normally distributed, which is often not the case. A safer bet is Mann-Whitney's U test, since this does not assume that the distributions are normal. For more information, see \cite{raschka2020model} and \cite{carrasco2020recent}. Also see \nameref{multcomparisons} and \nameref{significance}.

\subsection{Do correct for multiple comparisons} \label{multcomparisons}
Things get a bit more complicated when you want to use statistical tests to compare more than two models, since doing multiple pairwise tests is a bit like using the test set multiple times --- it can lead to overly-optimistic interpretations of significance. Basically, each time you carry out a comparison between two models using a statistical test, there's a probability that it will discover significant differences where there aren't any. This is represented by the confidence level of the test, usually set at 95\%: meaning that 1 in 20 times it will give you a false positive. For a single comparison, this may be a level of uncertainty you can live with. However, it accumulates. That is, if you do 20 pairwise tests with a confidence level of 95\%, one of them is likely to give you the wrong answer. This is known as the \textbf{multiplicity effect}, and is an example of a broader issue in data science known (at least when done intentionally) as \textbf{data dredging} or \textbf{p-hacking} --- see \cite{stefan2023big}. To address this problem, you can apply a correction for multiple tests. The most common approach is the Bonferroni correction, a very simple method that lowers the significance threshold based on the number of tests that are being carried out; see \cite{salzberg1997comparing} for a gentle introduction. However, there are numerous other approaches, and there is also some debate about when and where these corrections should be applied; for an accessible overview, see \cite{streiner2015best}.

\subsection{Don't always believe results from community benchmarks} \label{community}
In certain problem domains, it has become commonplace to use benchmark data sets to evaluate new ML models. The idea is that, because everyone is using the same data to train and test their models, then comparisons will be more transparent. Unfortunately this approach has some major drawbacks. First, if access to the test set is unrestricted, then you can't assume that people haven't used it as part of the training process. This is known as ``training to the test set'', and leads to results that are heavily over-optimistic. A more subtle problem is that, even if everyone who uses the data only uses the test set once, collectively the test set is being used many times by the community. In effect, by comparing lots of models on the same test set, it becomes increasingly likely that the best model just happens to over-fit the test set, and doesn't necessarily generalise any better than the other models (see \nameref{multcomparisons} and \nameref{validation}). For these, and other reasons, you should be careful how much you read into results from a benchmark data set, and don't assume that a small increase in performance is significant. This is particularly the case where foundation models (see \nameref{trends}) are used, since it is possible that their training data included the test sets from community benchmarks. See  \cite{paullada2021data} for a wider discussion of issues surrounding the use of shared datasets. Also see \nameref{measure}.

\subsection{Do combine models (carefully)} \label{ensemble}
Whilst this section focuses on comparing models, it's good to be aware that ML is not always about choosing between models. Often it makes sense to use combinations of models. Different ML models explore different trade-offs; by combining them, you can sometimes compensate for the weaknesses of one model by using the strengths of another model, and vice versa. \textbf{Ensembles} are a well-established group of composite models. There are lots of ensemble learning approaches --- see \cite{dong2020survey} for a review --- but they can be roughly divided into those that form ensembles out of the same base model type (examples include random forests, bagging and boosting) and those that combine different types of ML model. An example of the latter is \textbf{stacked generalisation} (or stacking), where a model is trained to aggregate the outputs of a group of base models. However, ensembles are not the only kind of composition. Another, increasingly common, form of composition occurs when \textbf{embedding models} (such as autoencoders or foundation models such as BERT) are used to provide input to other models. When using stacking or embedding, it's important to ensure that no data leaks (see \nameref{leakage}) occur, i.e.\ that the test data used to measure the performance of the composite model is not used in the training of any of its components. This is a common pitfall, especially when the model components are trained on overlapping data. To reduce the likelihood of sequential overfitting (see \nameref{validation}), it is also advisable to use a separate test set to evaluate the composite model.


\section{How to report your results} \label{report}
The aim of academic research is not self-aggrandisement, but rather an opportunity to contribute to knowledge. In order to effectively contribute to knowledge, you need to provide a complete picture of your work, covering both what worked and what didn't. ML is often about trade-offs --- it's very rare that one model is better than another in every way that matters --- and you should try to reflect this with a nuanced and considered approach to reporting results and conclusions.

\subsection{Do be transparent} \label{transparent}
First of all, always try to be transparent about what you've done, and what you've discovered, since this will make it easier for other people to build upon your work. In particular, it's good practice to share your models in an accessible way. For instance, if you used a script to implement all your experiments, then share the script when you publish the results. This means that other people can easily repeat your experiments, which adds confidence to your work. It also makes it a lot easier for people to compare models, since they no longer have to reimplement everything from scratch in order to ensure a fair comparison. Knowing that you will be sharing your work also encourages you to be more careful, document your experiments well, and write clean code, which benefits you as much as anyone else. It's also worth noting that issues surrounding reproducibility are gaining prominence in the ML community, so in the future you may not be able to publish work unless your workflow is adequately documented and shared --- for example, see \cite{pineau2021improving}. Checklists (\nameref{checklists}) are useful for knowing what to include in your workflow. You might also find \textbf{experiment tracking} frameworks, such as MLflow \citep{chen2020developments}, useful for recording your workflow.

\subsection{Do report performance in multiple ways} \label{measure}
One way to achieve better rigour when evaluating and comparing models is to use multiple data sets. This helps to overcome any deficiencies associated with individual data sets (see \nameref{community}) and allows you to present a more complete picture of your model's performance. It's also good practice to report multiple metrics for each data set, since different metrics can present different perspectives on the results, and increase the transparency of your work. For example, if you use accuracy, it's also a good idea to include metrics that are less sensitive to class imbalances (see \nameref{accuracy}). In domains such as medicine and security, it's important to know where errors are being made; for example, when your model gets things wrong, is it more inclined to false positives or false negatives? Metrics that summarise everything in one number, such as accuracy, give no insight into this. So, it's important to also include partial metrics such as precision and recall, or sensitivity and specificity, since these do provide insight into the types of errors your model produces. And make sure it's clear which metrics you are using. For instance, if you report F-scores, be clear whether this is $F_1$, or some other balance between precision and recall. If you report AUC, indicate whether this is the area under the ROC curve or the PR curve. For a broader discussion, see \cite{blagec2020critical}.

\subsection{Don't generalise beyond the data}\label{datalimits}
It's important not to present invalid conclusions, since this can lead other researchers astray. A common mistake is to make general statements that are not supported by the data used to train and evaluate models. For instance, if your model does really well on one data set, this does not mean that it will do well on other data sets. Whilst you can get more robust insights by using multiple data sets (see \nameref{measure}), there will always be a limit to what you can infer from any experimental study. There are numerous reasons for this (see \cite{paullada2021data}), many of which are to do with how datasets are curated. One common issue is bias, or \textbf{sampling error}: that the data is not sufficiently representative of the real world. Another is overlap: multiple data sets may not be independent, and may have similar biases. There's also the issue of quality: and this is a particular issue in deep learning datasets, where the need for quantity of data limits the amount of quality checking that can be done. So, in short, don't overplay your findings, and be aware of their limitations.

\subsection{Do be careful when reporting statistical significance} \label{significance}
I've already discussed statistical tests (see \nameref{statistical}), and how they can be used to determine differences between ML models. However, statistical tests are not perfect. Some are conservative, and tend to under-estimate significance; others are liberal, and tend to over-estimate significance. This means that a positive test doesn't always indicate that something is significant, and a negative test doesn't necessarily mean that something isn't significant. Then there's the issue of using a threshold to determine significance; for instance, a 95\% confidence threshold (i.e.\ when the p-value $<$ 0.05) means that 1 in 20 times a difference flagged as significant won't be significant. In fact, statisticians are increasingly arguing that it is better not to use thresholds, and instead just report p-values and leave it to the reader to interpret these \citep{betensky2019p}. Beyond statistical significance, another thing to consider is whether the difference between two models is actually important. If you have enough samples, you can always find significant differences, even when the actual difference in performance is miniscule. To give a better indication of whether something is important, you can measure \textbf{effect size}. There are a range of approaches used for this: Cohen's $d$ statistic is probably the most common, but more robust approaches, such as Kolmogorov-Smirnov, are preferable. For more on effect size and reporting statistical significance, see \cite{aguinis2021reporting}. You might also consider using Bayesian statistics; although there's less guidance and tools support available, these theoretically have a lot going for them, and they avoid many of the pitfalls associated with traditional statistical tests --- see \cite{benavoli2017time} for more info.

\subsection{Do look at your models} \label{look}

\begin{figure}[!tb]
	\centering
	\includegraphics[width=0.85\columnwidth]{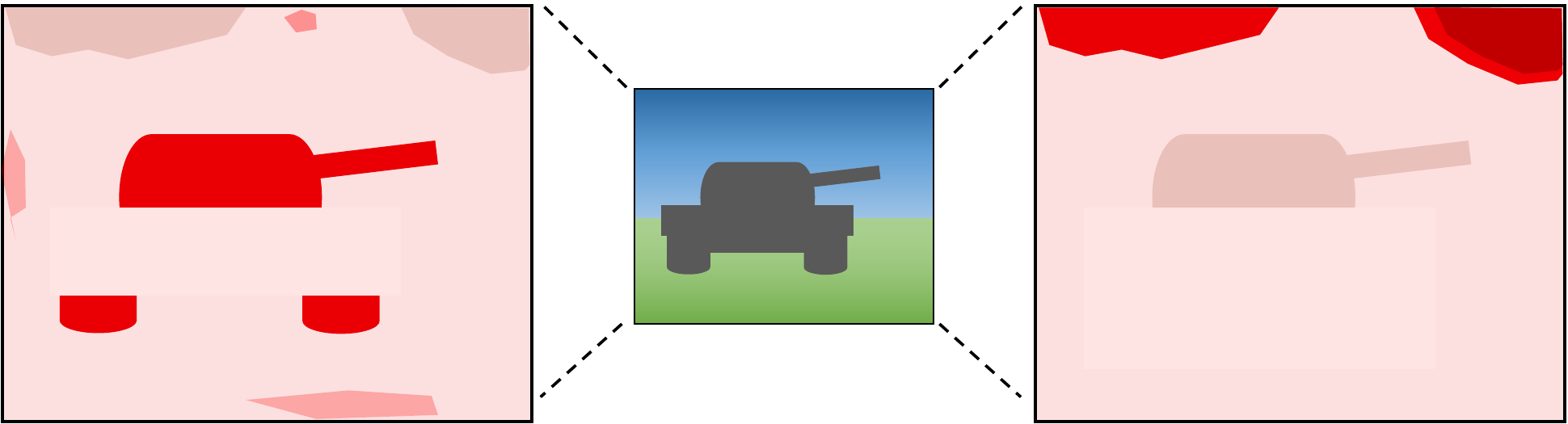}
	\caption{See \nameref{look}. Using saliency maps to analyse vision-based deep learning models. Imagine these two maps (in red) were generated for the image shown in the centre, for two different deep learning models trained on the kind of tank recognition data mentioned in \nameref{spurious}. Darker colours indicate features that are of greater importance to the model, so the model on the left (which predominantly focuses on the components of the tank) is likely to generalise much better than the one on the right (which predominantly focuses on the background of the image).}
	\label{fig:salientanks}
\end{figure}

Trained models contain a lot of useful information. Unfortunately many authors just report the performance metrics of a trained model, without giving any insight into what it actually learnt. Remember that the aim of research is not to get a slightly higher accuracy than everyone else. Rather, it's to generate knowledge and understanding and share this with the research community. If you can do this, then you're much more likely to get a decent publication out of your work. So, do look inside your models and do try to understand how they reach a decision. For relatively simple models like decision trees, it can also be beneficial to provide visualisations of your models, and most libraries have functions that will do this for you. For more complex models, there are a range of \textit{explainable AI} (XAI) techniques that can be used. Some of these are model-specific, and others are model-agnostic. Well-established examples of the latter are LIME and SHAP; both give insights into which features are important for a model. For CNNs and vision transformers, a common approach is to use \textbf{saliency maps}, which show the importance of different parts of an input image --- see Figure \ref{fig:salientanks} for an illustrative example. Grad-CAM is a popular technique for generating these, but there are plenty of other methods too. For non-vision transformers, a common approach is to visualise attention weights. See \cite{dwivedi2023explainable} for a survey of XAI techniques, and \cite{ali2023explainable} for a discussion of the limitations of current approaches. Whilst XAI techniques can give you useful insights into a model's behaviour, it's important to bear in mind that they are unlikely to tell you exactly what a model is doing. This is particularly the case for deep learning models (see \nameref{dnn}), whose complexity makes their behaviour inherently difficult to analyse. For complex models, ablation studies \cite{meyes2019ablation} can also be useful. This involves successively removing parts of the model to see what is important, and can result in a simpler model which is more amenable to analysis.

\subsection{Do use a machine learning checklist} \label{checklists}
This guide aims to give an appreciation of the main things that can go wrong during machine learning, plus some guidance on how to avoid these things going wrong. Checklists, on the other hand, are designed to take you more formally through the ML pipeline and encourage you to document (and more importantly, think about) how your implementation decisions support a meaningful outcome. In some domains, e.g.\ certain fields of medicine, it is compulsory to complete a checklist before submitting a paper for publication. However, beyond their quality assurance role, checklists are arguably most useful at the start of a study when it comes to planning an ML pipeline. Since I'm one of the authors, I'd particularly encourage you to look at \href{https://reforms.cs.princeton.edu}{REFORMS} \citep{kapoor2024reforms}, which is a combined checklist and set of consensus-based recommendations for doing ML-based science (although much of it is also applicable to ML practice more generally). Other, more domain-specific, checklists are also available.

\section{Final thoughts}
ML is becoming an important part of people's lives, yet the practice of ML is arguably in its infancy. There are many easy-to-make mistakes that can cause an ML model to appear to perform well, when in reality it does not. In turn, this has the potential to misinform when these models are published, and the potential to cause harm if these models are ever deployed. This guide describes the most common of these mistakes, and also touches upon more general issues of good practice in ML, such as fairness, transparency and the avoidance of bias. It also offers advice on avoiding these pitfalls. However, new threats continue to emerge as new approaches to ML are developed, and it is therefore important for users of ML to remain vigilant. This is the nature of a fast-moving research area --- the theory of how to do ML almost always lags behind the practice, practitioners will always disagree about the best ways of doing things, and what we think is correct today may not be correct tomorrow.
\vspace{4mm}

\noindent You can find more on ML pitfalls at my Substack, \href{https://fetchdecodeexecute.substack.com/}{Fetch Decode Execute}.

\section*{Acknowledgements}

Many thanks to everyone who gave me feedback on the draft manuscript, to everyone who has since sent me suggestions for revisions and new content, and to the editor and peer reviewers of the version published in \textit{Patterns}.

\renewcommand{\bibpreamble}{Where available, preprint URLs are also included for papers that are not open access.\vspace{2mm}}

\bibliographystyle{abbrvnat}
\bibliography{lones_arxiv}

\end{document}